\documentclass[
]{ceurart}

\usepackage{listings}
\begin{document}

\copyrightyear{2026}
\copyrightclause{Copyright for this paper by its authors.
  Use permitted under Creative Commons License Attribution 4.0
  International (CC BY 4.0).}

\conference{TRUST-AI: The Second European Workshop on Trustworthy AI. Organized as part of the International Joint Conference on Artificial Intelligence - IJCAI/ECAI 2026. August 2026, Bremen, Germany.}

\title{The First Token Knows: Single-Decode Confidence for Hallucination Detection}

\author[1]{Mina Gabriel}[%
    orcid=0009-0003-2058-3485,
    email=mina.gabriel@temple.edu,
    url=https://minagabriel.github.io/,
]
\cormark[1]
\cortext[1]{Corresponding author.}
\author[1]{Pei Wang}[%
    orcid=0000-0002-1066-0454,
    email=pei.wang@temple.edu,
    url=https://cis.temple.edu/~wangp/,
]
\address[1]{Department of Computer and Information Sciences, Temple University, Philadelphia, PA 19122, USA}

\begin{abstract}
Sampling-based uncertainty methods detect hallucinations by generating multiple answers and measuring agreement, often with additional semantic clustering. We study whether uncertainty is already visible in the first content-bearing answer token of a single greedy decode. We define first-token confidence, $\phi_{\mathrm{first}}$, as one minus the normalized entropy of the top-$K$ first-token distribution. Across PopQA and TriviaQA with three 7--8B instruction-tuned models, $\phi_{\mathrm{first}}$ matches or modestly exceeds semantic self-consistency while requiring only one decode instead of one greedy decode plus ten sampled generations and NLI clustering. A subsumption analysis shows that $\phi_{\mathrm{first}}$ is moderately to strongly correlated with semantic agreement, and combining both signals yields only small additional AUROC gains. We position $\phi_{\mathrm{first}}$ as a cheap deployment-oriented baseline for factual QA uncertainty estimation, while noting its limits for high-confidence falsehoods and long-form generation.
\end{abstract}

\maketitle

\section{Introduction}

A common paradigm for uncertainty quantification in large language models is \emph{self-consistency}: sample $N$ responses for the same input and use disagreement among them as a proxy for uncertainty. Originally proposed as a decoding strategy for reasoning~\cite{wang2022selfconsistency}, the same sampling-based principle has become central to several hallucination-detection methods. Semantic uncertainty refines this idea by clustering generations into NLI-based equivalence classes and treating disagreement among clusters as evidence of model uncertainty~\cite{kuhn2023semantic,lin2024generating}. These methods provide strong baselines, but require multiple generations per question
and a separate NLI-based clustering model. 

We argue that, in closed-book short-answer factual QA, where the model answers from its parametric knowledge without retrieved documents, sampling-based methods act
as expensive Monte Carlo probes of uncertainty that is already largely visible in
the model's first-token logit distribution. For factual questions such as
\emph{``Who wrote Hamlet?''} or \emph{``What is the capital of Australia?''},
the first generated answer token often marks the model's earliest commitment to
an entity, name, or relation value. If most of the probability mass at this
position is concentrated on one token, the model is making a confident early
choice about how to begin the answer. If the probability mass is instead spread
across several plausible first tokens, the model is unsure which answer to begin
generating, even before the rest of the response has unfolded.

We define first-token confidence $\phi_{\mathrm{first}}$ as the normalized
entropy of the top-$K$ logits at the first content-bearing answer token of a
single greedy decode and compare it against semantic self-consistency,
surface-form self-consistency, and verbalized confidence. We further test whether
$\phi_{\mathrm{first}}$ captures much of the same uncertainty information as
semantic agreement, which requires multiple sampled generations. 

Our contributions are: (i) we show that $\phi_{\mathrm{first}}$ matches or 
modestly exceeds semantic agreement on PopQA and TriviaQA across three 
7--8B models, at roughly $1/11$ of the generation cost, before accounting 
for the additional NLI clustering required by semantic agreement; (ii) we 
provide a subsumption test showing that $\phi_{\mathrm{first}}$ is moderately 
to strongly correlated with semantic agreement and that a logistic ensemble 
of the two adds only marginal AUROC over $\phi_{\mathrm{first}}$ alone; and 
(iii) we show that the apparent relationship between $\phi_{\mathrm{first}}$ 
and answer length is largely explained by correctness rather than answer 
length itself.\footnote{Code, per-question run outputs, and analysis scripts: 
\url{https://github.com/MinaGabriel/phi-signal}.}

Our use of the term hallucination detection is restricted to correctness prediction in closed-book factual QA, following common benchmark practice. We do not claim that first-token confidence detects all forms of hallucination. In particular, it is not expected to reliably identify high-confidence falsehoods, where the model assigns high probability to an incorrect answer. The proposed signal is therefore better viewed as a cheap uncertainty-based risk indicator: it can identify cases where the model's answer appears unstable or underdetermined, but it should be complemented by retrieval, external verification, or human review when confident errors are safety-critical.

\section{Method}

\subsection{First-token confidence}

Given a single greedy decode of a model's response, let $\ell_t \in \mathbb{R}^{|V|}$ denote the logits at decode step $t$ and $p_{t,i}$ the corresponding softmax probabilities. Let $t^\star$ be the position of the first content-bearing answer token, identified by skipping whitespace, punctuation, and chat-template prefixes such as ``Answer:''. We take the top-$K$ probabilities at position $t^\star$ (with $K=100$), renormalize them to $\tilde{p}_{t^\star,1}, \ldots, \tilde{p}_{t^\star,K}$, and define
\[
H_{t^\star} = -\sum_{i=1}^{K} \tilde{p}_{t^\star,i} \log \tilde{p}_{t^\star,i}, \qquad \phi_{\mathrm{first}} = 1 - \frac{H_{t^\star}}{\log K}.
\]
$\phi_{\mathrm{first}}$ ranges from $0$ (uniform top-$K$) to $1$ (all mass on a single token). It is computed from a single greedy forward pass: no additional sampling, no external models.

\subsection{Uncertainty baselines}

We sample $N = 10$ completions per question using temperature $0.7$ and top-$p = 0.95$. \textbf{AU-full} measures surface-form agreement by computing the fraction of sampled completions whose normalized full strings match the normalized greedy answer. \textbf{AU-3w} and \textbf{AU-1w} progressively relax this criterion to the first three words and the first word, providing increasingly strong surface-form baselines. \textbf{Semantic AU} performs meaning-level agreement by clustering the greedy answer and its $N$ samples using bidirectional NLI entailment with DeBERTa-v3-large-mnli~\cite{he2021deberta}, following the procedure of~\cite{kuhn2023semantic}, and reports the fraction of samples assigned to the greedy answer's cluster. \textbf{Verbalized confidence} prompts the model to output an integer from 0--100 reflecting its self-estimated correctness~\cite{tian2023just,xiong2023can}. We use the same sampling hyperparameters and scoring rules across all datasets and models. The resulting AUROC values should therefore be interpreted as untuned estimates rather than benchmark-specific optimized results.

\subsection{Cost}

$\phi_{\mathrm{first}}$ requires one greedy forward pass per question. Semantic AU
requires one greedy decode, $N=10$ sampled generations, and representative-based
bidirectional NLI clustering over the greedy and sampled answers. This requires
$O(CN)$ NLI comparisons, where $C$ is the number of discovered semantic clusters.
\section{Experiments}

\subsection{Setup}

We evaluate on the test split of PopQA~\cite{mallen2023when} and the validation split of TriviaQA~\cite{joshi2017triviaqa}, sampling $n=1000$ examples per dataset with a fixed seed. The same 1000 examples are used across all three models so that all comparisons are paired at the example level. We choose $n=1000$ as a compute--precision tradeoff. The standard error of an AUROC estimate decreases as $\Theta(1/\sqrt{n})$, so doubling to $n=2000$ would only narrow each cell's bootstrap interval by about $0.007$ AUROC points, while doubling all generation and NLI costs. We instead invest the saved compute in three models, two datasets, and the multi-method comparison, and report empirical 95\% bootstrap confidence intervals and paired bootstrap tests for every cell.

We evaluate three instruction-tuned 7--8B models: Llama-3.1-8B-Instruct~\cite{dubey2024llama}, Mistral-7B-Instruct-v0.3~\cite{jiang2023mistral}, and Qwen2.5-7B-Instruct~\cite{yang2024qwen2}. Correctness is determined by an automatic judge (Qwen2.5-14B-Instruct) given the
question, the model's answer, and gold aliases.

\subsection{Main results}

In this subsection, we compare $\phi_{\mathrm{first}}$ with verbalized
confidence, surface-form self-consistency, and semantic self-consistency. The
main question is whether a single-decode token-level confidence signal can match
or exceed uncertainty signals that require multiple sampled generations.

\begin{figure*}[t]
\centering
\includegraphics[width=\textwidth]{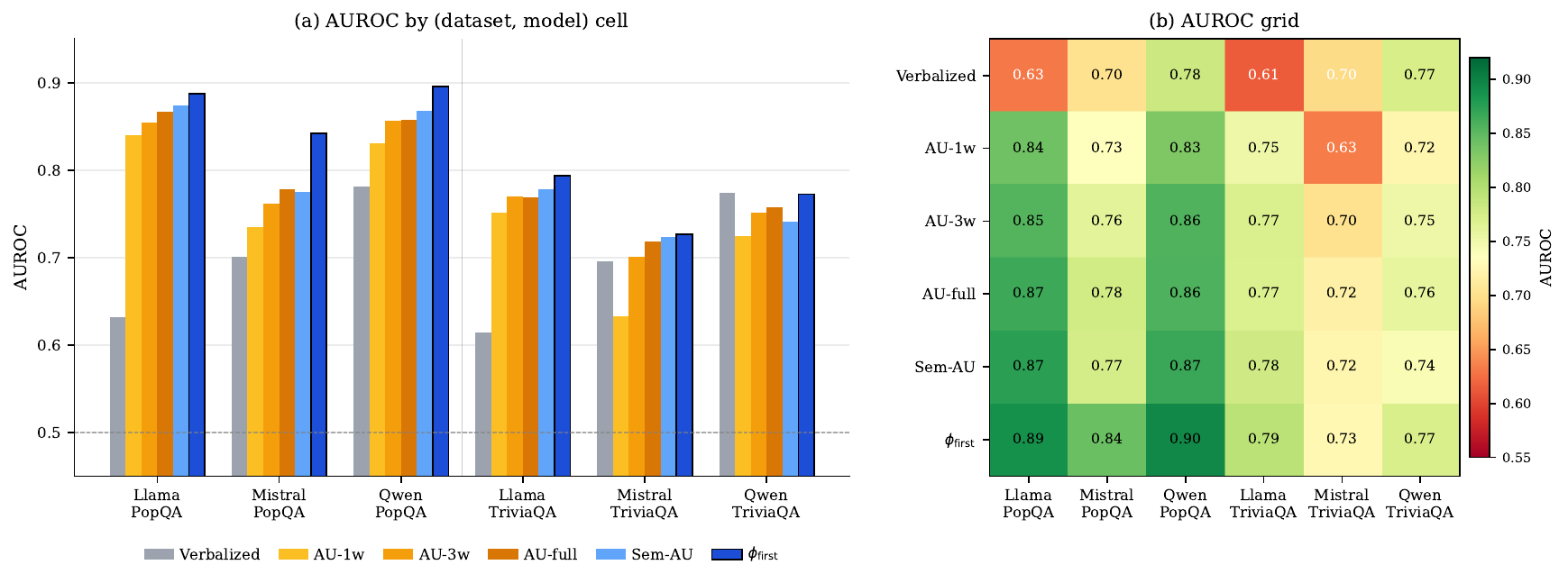}
\caption{AUROC of correctness prediction across six dataset--model cells and six confidence signals ($n{=}1000$ per cell). \textbf{(a)} Grouped bars show AUROC per cell; the dashed line marks chance performance. \textbf{(b)} The same values are shown as a heatmap. $\phi_{\mathrm{first}}$ achieves the highest AUROC in five of six cells and is within $0.002$ AUROC of the strongest method in the remaining cell, while requiring only a single greedy decode.}
\label{fig:auroc-overview}
\end{figure*}

\begin{table*}[h]
\centering
\small
\setlength{\tabcolsep}{5pt}
\renewcommand{\arraystretch}{1.15}
\caption{AUROC for correctness prediction in closed-book factual QA across three 7--8B instruction-tuned models on PopQA and TriviaQA ($n=1000$ each). Methods are grouped by inference cost. Best per row in \textbf{bold}; second-best \underline{underlined}.
$\Delta$ is the gap between $\phi_{\mathrm{first}}$ and the strongest
non-$\phi$ baseline.}

\label{tab:main}
\begin{tabular}{llc cccc c@{\hskip 8pt}c@{\hskip 6pt}c}
\toprule
& & \multicolumn{1}{c}{\textbf{1 decode}} & \multicolumn{4}{c}{\textbf{$1{+}10$ decodes (sampling)}} & \multicolumn{1}{c}{\textbf{1 decode}} & & \\
\cmidrule(lr){3-3} \cmidrule(lr){4-7} \cmidrule(lr){8-8}
\textbf{Dataset} & \textbf{Model} & \textbf{Verb.} & \textbf{AU-1w} & \textbf{AU-3w} & \textbf{AU-full} & \textbf{Sem.\ AU} & $\boldsymbol{\phi_{\mathrm{first}}}$ & $\boldsymbol{\Delta}$ & \textbf{Acc.} \\
\midrule
\multirow{3}{*}{\textbf{PopQA}}
 & Llama-3.1-8B   & 0.632 & 0.840 & 0.854 & 0.866 & \underline{0.874} & \textbf{0.887} & $+$0.013 & 0.27 \\
 & Mistral-7B-v0.3& 0.701 & 0.735 & 0.762 & \underline{0.778} & 0.775 & \textbf{0.842} & $+$0.064 & 0.25 \\
 & Qwen2.5-7B     & 0.782 & 0.831 & 0.856 & 0.857 & \underline{0.867} & \textbf{0.895} & $+$0.028 & 0.19 \\
\cmidrule(l){1-10}
\multirow{3}{*}{\textbf{TriviaQA}}
 & Llama-3.1-8B   & 0.614 & 0.752 & 0.770 & 0.769 & \underline{0.778} & \textbf{0.794} & $+$0.016 & 0.64 \\
 & Mistral-7B-v0.3& 0.696 & 0.632 & 0.701 & 0.718 & \underline{0.724} & \textbf{0.727} & $+$0.003 & 0.62 \\
 & Qwen2.5-7B     & \textbf{0.774} & 0.725 & 0.751 & 0.758 & 0.741 & \underline{0.772} & $-$0.002 & 0.52 \\
\cmidrule(l){1-10}
\rowcolor{gray!8}
\multicolumn{2}{l}{\textit{PopQA mean}}    & 0.705 & 0.802 & 0.824 & 0.834 & 0.839 & \textbf{0.875} & $+$0.036 & 0.24 \\
\rowcolor{gray!8}
\multicolumn{2}{l}{\textit{TriviaQA mean}} & 0.695 & 0.703 & 0.741 & 0.748 & 0.748 & \textbf{0.764} & $+$0.016 & 0.59 \\
\midrule
\rowcolor{gray!12}
\multicolumn{2}{l}{\textbf{Overall mean}}  & 0.700 & 0.752 & 0.782 & 0.791 & 0.793 & \textbf{0.820} & $+$0.027 & 0.42 \\
\bottomrule
\end{tabular}
\end{table*}

Figure~\ref{fig:auroc-overview} summarizes our main result visually, and
Table~\ref{tab:main} reports the corresponding numbers. Panel (a) of
Figure~\ref{fig:auroc-overview} shows AUROC as grouped bars per
dataset--model cell, with $\phi_{\mathrm{first}}$ highlighted; panel (b)
presents the same values as a heatmap for at-a-glance comparison across
methods. Both views show the same pattern: $\phi_{\mathrm{first}}$ is
the strongest method in five of six dataset--model cells and is within
$0.002$ AUROC of the strongest method in the remaining cell. The pattern
is consistent across both datasets: $\phi_{\mathrm{first}}$ improves the
per-dataset mean by $+0.036$ AUROC on PopQA ($0.875$ vs.\ $0.839$ for
semantic AU) and by $+0.016$ on TriviaQA ($0.764$ vs.\ $0.748$). The
smaller TriviaQA gain suggests that longer and more lexically variable
answers give sampling-based methods relatively more opportunity to recover
useful agreement information; we return to this point in the limitations.

In the overall mean, $\phi_{\mathrm{first}}$ reaches $0.820$ AUROC, compared with
$0.793$ for semantic AU, $0.791$ for AU-full, $0.782$ for AU-3w, and $0.752$ for
AU-1w. Verbalized confidence is weaker, with a mean AUROC of $0.700$, consistent
with prior work showing that LLMs are often poorly calibrated when asked to state
their own confidence directly~\cite{tian2023just}. Thus, the advantage of
$\phi_{\mathrm{first}}$ over semantic AU is modest in absolute terms
($+2.7$ AUROC points on average), but it is obtained with a single greedy decode
rather than multiple sampled generations and NLI-based semantic clustering.

\subsection{Statistical reliability of the gains}
The AUROC results show the size of the performance differences, but do not
show by themselves whether those differences are stable between evaluation
examples. We therefore use paired bootstrap resampling over questions to compare
$\phi_{\mathrm{first}}$ against the main baselines within each dataset--model
cell. Because both methods are evaluated on the same questions, the test measures
whether the observed AUROC gap is robust to resampling of the evaluation set.
\begin{table}[h]
\centering
\footnotesize
\setlength{\tabcolsep}{5pt}
\renewcommand{\arraystretch}{1.15}
\caption{Paired bootstrap test of $\Delta\mathrm{AUROC} > 0$ for $\phi_{\mathrm{first}}$ vs.\ each baseline ($B{=}1000$ resamples; one-sided $p$). With $B=1000$, the smallest resolvable $p$-value is $\approx 0.001$, so cells reported as $<0.001$ correspond to $0$/$1000$ resamples favoring the baseline. \textbf{Bold} indicates $p < 0.05$.}
\label{tab:bootstrap}
\begin{tabular}{llccc}
\toprule
\textbf{Dataset} & \textbf{Model} & \textbf{vs.\ AU-full} & \textbf{vs.\ Sem.\ AU} & \textbf{vs.\ AU-1w} \\
\midrule
\multirow{3}{*}{\textbf{PopQA}}
 & Llama   & \textbf{0.018} & 0.082 & \textbf{$<$0.001} \\
 & Mistral & \textbf{$<$0.001} & \textbf{$<$0.001} & \textbf{$<$0.001} \\
 & Qwen    & \textbf{0.001} & \textbf{0.008} & \textbf{$<$0.001} \\
\cmidrule(l){1-5}
\multirow{3}{*}{\textbf{TriviaQA}}
 & Llama   & \textbf{0.025} & 0.105 & \textbf{$<$0.001} \\
 & Mistral & 0.301 & 0.416 & \textbf{$<$0.001} \\
 & Qwen    & 0.153 & \textbf{0.014} & \textbf{$<$0.001} \\
\midrule
\rowcolor{gray!12}
\multicolumn{2}{l}{Significant ($p<0.05$)} & 4/6 & 3/6 & 6/6 \\
\bottomrule
\end{tabular}
\end{table}

Table~\ref{tab:bootstrap} reports paired bootstrap tests over questions. These
tests ask whether the AUROC advantage of $\phi_{\mathrm{first}}$ over each
baseline is stable under resampling of the same evaluation examples. The results show that $\phi_{\mathrm{first}}$ significantly outperforms AU-full
in four of six cells and semantic AU in three of six cells. The remaining
semantic-AU differences are not statistically significant, so we frame
$\phi_{\mathrm{first}}$ as \emph{matching} semantic self-consistency rather than
uniformly outperforming it. Against AU-1w, the simplest surface-form baseline,
the gain is significant in all six cells.

\subsection{Subsumption analysis}

We test whether $\phi_{\mathrm{first}}$ already captures the information provided by semantic AU. For each cell we report two quantities: the Pearson correlation between $\phi_{\mathrm{first}}$ and semantic AU, and the AUROC gain obtained by combining both signals in a standardized logistic regression over $\phi_{\mathrm{first}}$ alone. A high correlation paired with a near-zero ensemble gain indicates that semantic AU adds little beyond $\phi_{\mathrm{first}}$.

\begin{table}[h]
\centering
\footnotesize
\setlength{\tabcolsep}{6pt}
\renewcommand{\arraystretch}{1.15}
\caption{Subsumption analysis. \textbf{Pearson $r$}: correlation between $\phi_{\mathrm{first}}$ and semantic AU. \textbf{Gain}: AUROC of the logistic ensemble of both signals minus AUROC of $\phi_{\mathrm{first}}$ alone. Small gains indicate semantic AU adds little beyond $\phi_{\mathrm{first}}$.}
\label{tab:subsumption}
\begin{tabular}{llcc}
\toprule
\textbf{Dataset} & \textbf{Model} & \textbf{Pearson $r$} & \textbf{Ensemble gain} \\
\midrule
\multirow{3}{*}{\textbf{PopQA}}
 & Llama   & 0.76 & $+$0.017 \\
 & Mistral & 0.59 & $+$0.009 \\
 & Qwen    & 0.75 & $+$0.012 \\
\cmidrule(l){1-4}
\multirow{3}{*}{\textbf{TriviaQA}}
 & Llama   & 0.74 & $+$0.019 \\
 & Mistral & 0.54 & $+$0.045 \\
 & Qwen    & 0.67 & $+$0.024 \\
\midrule
\rowcolor{gray!12}
\multicolumn{2}{l}{\textbf{Mean}} & \textbf{0.67} & $\boldsymbol{+0.021}$ \\
\bottomrule
\end{tabular}
\end{table}

Three observations follow from Table~\ref{tab:subsumption}. First, $\phi_{\mathrm{first}}$ and semantic AU are moderately to strongly correlated, with Pearson $r$ between $0.54$ and $0.76$ (mean $0.67$). Second, combining the two signals improves AUROC by only $+0.021$ on average, and by less than $+0.025$ in five of six cells. Third, $\phi_{\mathrm{first}}$ alone matches or exceeds semantic AU's standalone AUROC in every cell, so the residual ensemble gain reflects a small complementary contribution from semantic AU rather than a deficit in $\phi_{\mathrm{first}}$. Together, these results indicate that $\phi_{\mathrm{first}}$ already captures most of the discriminative content that semantic agreement extracts at substantially higher inference cost.

\subsection{Length confound}

A natural concern is that $\phi_{\mathrm{first}}$ may simply track the length of the generated answer. We test this in two stages. First, we compute the raw Pearson correlation $r_{\mathrm{len}}$ between $\phi_{\mathrm{first}}$ and the number of generated answer tokens. Second, since wrong answers tend to be both longer and lower-confidence, we control for correctness by computing the partial Pearson correlation $r_{\mathrm{len}}^{\mathrm{partial}}$ between $\phi_{\mathrm{first}}$ and answer length after removing the linear effect of the binary correctness label from both variables.

\begin{table}[h]
\centering
\footnotesize
\setlength{\tabcolsep}{6pt}
\renewcommand{\arraystretch}{1.15}
\caption{Length confound. \textbf{$r_{\mathrm{len}}$}: raw Pearson correlation between $\phi_{\mathrm{first}}$ and answer length. \textbf{$r_{\mathrm{len}}^{\mathrm{partial}}$}: partial correlation controlling for correctness. Values close to zero after partialling indicate that length is not driving $\phi_{\mathrm{first}}$.}
\label{tab:length}
\begin{tabular}{llcc}
\toprule
\textbf{Dataset} & \textbf{Model} & \textbf{$r_{\mathrm{len}}$} & \textbf{$r_{\mathrm{len}}^{\mathrm{partial}}$} \\
\midrule
\multirow{3}{*}{\textbf{PopQA}}
 & Llama   & $-$0.16 & $-$0.02 \\
 & Mistral & $-$0.13 & $-$0.03 \\
 & Qwen    & $-$0.14 & $-$0.04 \\
\cmidrule(l){1-4}
\multirow{3}{*}{\textbf{TriviaQA}}
 & Llama   & $-$0.23 & $-$0.18 \\
 & Mistral & $-$0.25 & $-$0.17 \\
 & Qwen    & $-$0.11 & $-$0.05 \\
\bottomrule
\end{tabular}
\end{table}
Table~\ref{tab:length} reports both quantities. The raw correlation ranges from $-0.11$ to $-0.25$ across cells, accounting for at most $6.5\%$ of the variance in $\phi_{\mathrm{first}}$. On PopQA, the partial correlation shrinks substantially: from $-0.16$ to $-0.02$ for Llama and from $-0.13$ to $-0.03$ for Mistral. This suggests that the apparent length effect on PopQA is largely explained by correctness rather than answer length itself. On TriviaQA, the partial correlation drops by less: a residual correlation of about $-0.18$ remains for Llama and Mistral. This indicates a small but non-trivial residual sensitivity to answer length on TriviaQA, which we list as a limitation.

\section{Related work}

Semantic self-consistency~\cite{kuhn2023semantic,lin2024generating} estimates uncertainty from disagreement among NLI-based equivalence classes of sampled generations. Surface-form variants instead measure agreement over normalized strings or answer prefixes. Other single-pass approaches include token probabilities, sequence-level likelihood~\cite{malinin2021uncertainty}, internal probes~\cite{kadavath2022language,azaria2023internal}, and verbalized confidence~\cite{tian2023just,xiong2023can}. Our work directly evaluates first-token entropy as a standalone signal against semantic self-consistency and tests how much semantic-agreement information it captures.

\section{Implications for responsible LLM deployment}

Because $\phi_{\mathrm{first}}$ requires only one greedy decode, it can be logged or monitored in deployed factual QA systems at low cost. Low-confidence answers can trigger abstention, retrieval, verification, or human review before being shown to users.

The score should not be treated as a guarantee of correctness. Thresholds must be calibrated for the target domain, and systems should report error rates and coverage at different operating points. In high-stakes settings, $\phi_{\mathrm{first}}$ is best used as an inexpensive first-stage signal within a broader assurance pipeline.

\section{Discussion and conclusion}

First-token confidence matches or modestly exceeds semantic self-consistency on closed-book factual QA across three 7--8B instruction-tuned models, while requiring roughly $1/11$ of the generation cost. The subsumption analysis shows that $\phi_{\mathrm{first}}$ is moderately to strongly correlated with semantic agreement and recovers much of its discriminative content from a single greedy decode. We recommend reporting $\phi_{\mathrm{first}}$ as a cheap baseline before claiming gains from sampling-based uncertainty methods.

\paragraph{Limitations.} This study is limited to English closed-book short-answer factual QA with three open 7--8B models and two benchmarks. Results may not transfer to long-form generation, retrieval-augmented QA, multilingual settings, larger or proprietary models, or black-box APIs without token probabilities. We do not evaluate all single-pass baselines, such as sequence likelihood or P(True)-style prompting. We fix $K=100$ rather than sweeping $K$, and use Qwen2.5-14B-Instruct as the judge, which may introduce model-family bias for the Qwen generator. Future work should test $K$ sensitivity, additional baselines, non-Qwen judges, and human annotations.

\section{Ethical Considerations}

First-token confidence is not a complete hallucination detector. It is a low-cost uncertainty signal for closed-book factual QA and can help flag answers that need abstention, retrieval, verification, or human review. High confidence does not guarantee correctness, especially for high-confidence falsehoods.

For responsible deployment, $\phi_{\mathrm{first}}$ should be used as one component of a broader reliability pipeline, not as a standalone safety mechanism. In high-stakes settings, it should be combined with external grounding, domain-specific evaluation, calibrated thresholds, and human oversight.

\section*{Declaration on Generative AI}

During the preparation of this work, the author(s) used OpenAI ChatGPT for language editing, LaTeX formatting assistance, and revision planning. After using this tool, the author(s) reviewed and edited the content as needed and take full responsibility for the publication's content.

\bibliography{sample-ceur}

@inproceedings{wang2022selfconsistency,
  title     = {Self-Consistency Improves Chain of Thought Reasoning in Language Models},
  author    = {Wang, Xuezhi and Wei, Jason and Schuurmans, Dale and Le, Quoc and Chi, Ed and Narang, Sharan and Chowdhery, Aakanksha and Zhou, Denny},
  booktitle = {International Conference on Learning Representations (ICLR)},
  year      = {2023}
}

@inproceedings{kuhn2023semantic,
  title     = {Semantic Uncertainty: Linguistic Invariances for Uncertainty Estimation in Natural Language Generation},
  author    = {Kuhn, Lorenz and Gal, Yarin and Farquhar, Sebastian},
  booktitle = {International Conference on Learning Representations (ICLR)},
  year      = {2023}
}

@article{lin2024generating,
  title   = {Generating with Confidence: Uncertainty Quantification for Black-box Large Language Models},
  author  = {Lin, Zhen and Trivedi, Shubhendu and Sun, Jimeng},
  journal = {Transactions on Machine Learning Research},
  year    = {2024}
}

@inproceedings{he2021deberta,
  title     = {{DeBERTa}: Decoding-enhanced {BERT} with Disentangled Attention},
  author    = {He, Pengcheng and Liu, Xiaodong and Gao, Jianfeng and Chen, Weizhu},
  booktitle = {International Conference on Learning Representations (ICLR)},
  year      = {2021}
}

@inproceedings{mallen2023when,
  title     = {When Not to Trust Language Models: Investigating Effectiveness of Parametric and Non-Parametric Memories},
  author    = {Mallen, Alex and Asai, Akari and Zhong, Victor and Das, Rajarshi and Khashabi, Daniel and Hajishirzi, Hannaneh},
  booktitle = {Proceedings of the 61st Annual Meeting of the Association for Computational Linguistics (ACL)},
  year      = {2023}
}

@inproceedings{joshi2017triviaqa,
  title     = {{TriviaQA}: A Large Scale Distantly Supervised Challenge Dataset for Reading Comprehension},
  author    = {Joshi, Mandar and Choi, Eunsol and Weld, Daniel S. and Zettlemoyer, Luke},
  booktitle = {Proceedings of the 55th Annual Meeting of the Association for Computational Linguistics (ACL)},
  year      = {2017}
}

@article{dubey2024llama,
  title   = {The {Llama} 3 Herd of Models},
  author  = {Dubey, Abhimanyu and others},
  journal = {arXiv preprint arXiv:2407.21783},
  year    = {2024}
}

@article{jiang2023mistral,
  title   = {{Mistral 7B}},
  author  = {Jiang, Albert Q. and Sablayrolles, Alexandre and Mensch, Arthur and Bamford, Chris and Chaplot, Devendra Singh and de las Casas, Diego and Bressand, Florian and Lengyel, Gianna and Lample, Guillaume and Saulnier, Lucile and others},
  journal = {arXiv preprint arXiv:2310.06825},
  year    = {2023}
}

@article{yang2024qwen2,
  title   = {{Qwen2.5} Technical Report},
  author  = {Yang, An and others},
  journal = {arXiv preprint arXiv:2412.15115},
  year    = {2024}
}

@inproceedings{tian2023just,
  title     = {Just Ask for Calibration: Strategies for Eliciting Calibrated Confidence Scores from Language Models Fine-Tuned with Human Feedback},
  author    = {Tian, Katherine and Mitchell, Eric and Zhou, Allan and Sharma, Archit and Rafailov, Rafael and Yao, Huaxiu and Finn, Chelsea and Manning, Christopher D.},
  booktitle = {Proceedings of the 2023 Conference on Empirical Methods in Natural Language Processing (EMNLP)},
  year      = {2023}
}

@article{xiong2023can,
  title   = {Can {LLMs} Express Their Uncertainty? An Empirical Evaluation of Confidence Elicitation in {LLMs}},
  author  = {Xiong, Miao and Hu, Zhiyuan and Lu, Xinyang and Li, Yifei and Fu, Jie and He, Junxian and Hooi, Bryan},
  journal = {arXiv preprint arXiv:2306.13063},
  year    = {2023}
}

@inproceedings{malinin2021uncertainty,
  title     = {Uncertainty Estimation in Autoregressive Structured Prediction},
  author    = {Malinin, Andrey and Gales, Mark},
  booktitle = {International Conference on Learning Representations (ICLR)},
  year      = {2021}
}

@article{kadavath2022language,
  title   = {Language Models (Mostly) Know What They Know},
  author  = {Kadavath, Saurav and Conerly, Tom and Askell, Amanda and others},
  journal = {arXiv preprint arXiv:2207.05221},
  year    = {2022}
}

@inproceedings{azaria2023internal,
  title     = {The Internal State of an {LLM} Knows When It's Lying},
  author    = {Azaria, Amos and Mitchell, Tom},
  booktitle = {Findings of the Association for Computational Linguistics: EMNLP},
  year      = {2023}
}

\appendix


\section{Experimental Details}
\label{app:experimental-details}

This appendix summarizes the key implementation details needed to reproduce
the experiments. All code, prompts, and run scripts are available at 
\url{https://github.com/MinaGabriel/phi-signal}.

\subsection{Hardware and software}
\label{app:hardware}

All experiments were run on a single NVIDIA A100 40\,GB GPU. Models were
loaded in \texttt{bfloat16}. We used \texttt{transformers},
\texttt{accelerate}, and \texttt{datasets}; all runs used random seed $42$.

\subsection{Models and checkpoints}
\label{app:models}

\begin{table}[h]
\centering
\footnotesize
\setlength{\tabcolsep}{6pt}
\renewcommand{\arraystretch}{1.15}
\caption{Model checkpoints used. All models were loaded in \texttt{bfloat16}.}
\label{tab:models}
\begin{tabular}{ll}
\toprule
\textbf{Role} & \textbf{HuggingFace path} \\
\midrule
Generator (Llama)   & \texttt{meta-llama/Meta-Llama-3.1-8B-Instruct} \\
Generator (Mistral) & \texttt{mistralai/Mistral-7B-Instruct-v0.3} \\
Generator (Qwen)    & \texttt{Qwen/Qwen2.5-7B-Instruct} \\
Judge               & \texttt{Qwen/Qwen2.5-14B-Instruct} \\
NLI clustering      & \texttt{MoritzLaurer/DeBERTa-v3-large-mnli-fever-anli-ling-wanli} \\
\bottomrule
\end{tabular}
\end{table}

\subsection{Datasets}
\label{app:datasets}

We used \texttt{akariasai/PopQA} and \texttt{mandarjoshi/trivia\_qa}
with the \texttt{rc.nocontext} configuration. For each dataset, we used
$n=1000$ examples in a fixed deterministic order and reused the same
indices across all generator models, making all comparisons paired at
the example level.

\subsection{Generation settings}
\label{app:gen-hyperparams}

\begin{table}[h]
\centering
\footnotesize
\setlength{\tabcolsep}{6pt}
\renewcommand{\arraystretch}{1.15}
\caption{Generation settings used across datasets and generator models.}
\label{tab:gen-hyperparams}
\begin{tabular}{ll}
\toprule
\textbf{Parameter} & \textbf{Value} \\
\midrule
Greedy max new tokens          & 32 \\
Sample max new tokens          & 32 \\
Number of samples per question & $N = 10$ \\
Sampling temperature           & $0.7$ \\
Sampling top-$p$               & $0.95$ \\
Top-$K$ for $\phi_{\mathrm{first}}$ & $K = 100$ \\
Verbalized confidence max new tokens & 6 \\
Random seed                    & $42$ \\
\bottomrule
\end{tabular}
\end{table}

\subsection{Prompts}
\label{app:prompts}

For all generator models, we used each model's native chat template via
\texttt{tokenizer.apply\_chat\_template} with
\texttt{add\_generation\_prompt=True}.

\paragraph{Answering prompt.}

\begin{quote}\small
\textbf{System:} \emph{You are a concise factual QA assistant. Answer in
as few words as possible. No explanations.}\\[2pt]
\textbf{User:} \emph{Question:} \texttt{\{question\}}\\
\emph{Answer:}
\end{quote}

\paragraph{Verbalized confidence prompt.}

\begin{quote}\small
\textbf{System:} \emph{You are calibrating the confidence of an answer.
Reply with a single integer from 0 to 100 representing the probability
that the answer is correct. No words.}\\[2pt]
\textbf{User:} \emph{Question:} \texttt{\{question\}}\\
\emph{Answer:} \texttt{\{model\_answer\}}\\
\emph{Confidence (0--100):}
\end{quote}

The first integer in the model's reply, clipped to $[0,100]$, is divided
by 100 and used as the verbalized confidence score.

\paragraph{Judge prompt.}

\begin{quote}\small
\textbf{System:} \emph{You are a strict factual grading assistant.
Decide if the model's answer is factually correct given the question
and gold answers. Respond with exactly one token: YES or NO.}\\[2pt]
\textbf{User:} \emph{Question:} \texttt{\{question\}}\\
\emph{Model answer:} \texttt{\{model\_answer\}}\\
\emph{Acceptable gold answers:} \texttt{\{gold\_list\}}\\
\emph{Is the model answer correct? YES or NO.}
\end{quote}

We greedy-decode up to 4 new tokens and set \texttt{is\_correct} $=1$
iff the decoded text begins with \texttt{YES}, case-insensitive.

\subsection{First-token score}
\label{app:first-token}

The first content-bearing answer token $t^\star$ is found by scanning
the greedy decoded tokens and skipping leading punctuation, whitespace,
and literal answer prefixes such as \texttt{Answer}. If no such token is
found, we use the first decoded token. We then compute
$\phi_{\mathrm{first}} = 1 - H_{t^\star}/\log K$ using the renormalized
top-$K$ distribution at $t^\star$, with $K=100$.

\subsection{Agreement baselines}
\label{app:agreement-baselines}

For surface-form agreement, sampled and greedy answers are normalized by
lower-casing, removing articles, removing punctuation, and collapsing
whitespace. \textbf{AU-full} compares the full normalized answer,
\textbf{AU-3w} compares the first three words, and \textbf{AU-1w}
compares the first word. Each score is the fraction of the $N=10$
samples matching the greedy answer under that key.

For semantic agreement, we cluster the greedy answer and its samples
using bidirectional NLI entailment following~\cite{kuhn2023semantic}.
Semantic AU is the fraction of samples assigned to the greedy answer's
semantic cluster.

\subsection{Statistical procedures}
\label{app:stats}

For each model--dataset cell, we compute AUROC over the same $n=1000$
questions. Confidence intervals and paired comparisons are estimated
with $B=1000$ bootstrap resamples over question indices.

For the subsumption experiment, we fit a standardized logistic regression
on $(\phi_{\mathrm{first}}, \mathrm{Sem.\ AU})$ within each cell and
compare its AUROC to $\phi_{\mathrm{first}}$ alone. For the length
analysis, we report the partial Pearson correlation between
$\phi_{\mathrm{first}}$ and answer length after residualizing both
variables against correctness.

\end{document}